\documentclass[runningheads]{llncs}

\usepackage[T1]{fontenc}
\usepackage{newtxtext}
\usepackage[varvw]{newtxmath}
\usepackage{graphicx}
\usepackage{url}

\begin{document}

\title{When Valid Decisions Are Not Enough:
Evaluating Structural Constraints in
Knowledge-Guided Generative Systems}
\titlerunning{Structural Constraints in Generative Systems}

% Author version for arXiv (not for double-blind conference submission).
\author{Josef Pavl\'{\i}\v{c}ek\inst{1}\orcidID{0000-0002-3959-5406}
\and Petra Pavl\'{\i}\v{c}kov\'{a}\inst{1}\orcidID{0000-0002-7456-2108}
\and Martin Molhanec\inst{2}\orcidID{0000-0003-2032-1138}}
\authorrunning{J. Pavl\'{\i}\v{c}ek et al.}
\institute{Department of Software Engineering, Faculty of Information Technology,
Czech Technical University in Prague, Czech Republic\\
\email{Josef.Pavlicek@fit.cvut.cz, Petra.Pavlickova@fit.cvut.cz}
\and Faculty of Electrical Engineering, Czech Technical University in Prague,
Prague 6, Czech Republic\\
\email{molhanec@cvut.cz}}

\maketitle

\begin{abstract}
Knowledge-guided generative systems integrate explicit domain
constraints to improve decision consistency, interpretability,
and alignment with human-created solutions. However, locally
valid decisions and structurally consistent outputs do not
necessarily preserve the contextual relationships observed in
human decision-making. This paper investigates how explicit
structural constraints influence the behavior of
optimization-based generative systems and under which conditions
they improve or degrade agreement with human reference solutions.
We introduce HARMONIX, an experimental framework that uses
automated musical harmonization as a controlled testbed for
studying general engineering challenges in knowledge-guided
decision-making. Four configurations sharing an identical
Viterbi-based optimization mechanism are evaluated: an
unconstrained structural baseline, repetition constraints,
functional constraints, and their combination. The experiments
comprise 19,745 records and 143,634 motif pairs across training
and validation partitions. Results demonstrate a consistent
context-dependent trade-off: combined constraints improve
reference interval agreement by approximately five percentage
points when structural relationships remain stable, but reduce
agreement by nearly nine percentage points when reference
solutions follow contextual transformations. These effects are
reproduced across both evaluation partitions, whereas changes
in overall exact decision agreement remain marginal. Contexts
are identified retrospectively using reference annotations;
we evaluate constraint effects rather than propose an adaptive
constraint-selection mechanism. The
findings show that adding interpretable constraints does not
necessarily improve agreement with reference solutions and that aggregate
evaluation metrics can conceal systematic context-specific
failures. From an information systems engineering perspective,
the study provides empirical evidence for the limitations of
static constraint enforcement and identifies context-aware
constraint applicability as an important design consideration
for knowledge-guided generative systems.

\end{abstract}

\section{Introduction}
\label{sec:introduction}

Knowledge-guided generative systems combine computational optimization with explicit domain knowledge. Constraints can improve consistency and transparency, yet a meaningful rule need not be appropriate in every decision context~\cite{bonlarron_constraining_2026}. This distinction matters when human reference solutions preserve some repeated relationships but deliberately transform others. A fixed preference may therefore increase agreement in one context and reduce it in another, while aggregate metrics obscure the difference.

Context is not merely an additional input variable: it determines which information is relevant to a decision and how that information should be interpreted~\cite{dey_understanding_2001}. In information systems engineering, this issue arises when decision support must respond to changing process conditions. Yurt et al.~\cite{yurt_guiding_2022} examine guidance for knowledge workers under dynamic contexts, while Leemans and Tax~\cite{leemans_causal_2022} investigate causal relationships among control-flow decisions. These studies concern business processes rather than generative harmonization, but illustrate why decision relationships and their conditions deserve explicit analysis.

Explainable AI emphasizes transparent decisions~\cite{adadi_peeking_2018,barredo_arrieta_explainable_2020}. Methods for explaining individual predictions, such as those examined by Ribeiro et al.~\cite{ribeiro_why_2016}, and broader surveys of black-box explanation~\cite{guidotti_survey_2018} help characterize what an explanation communicates. However, being able to explain why a preference influenced an output does not establish that the preference was appropriate for the context. Human--AI interaction guidelines further motivate examining how computational guidance relates to human decision-making~\cite{amershi_guidelines_2019}. At the information-systems level, Kubrak et al.~\cite{kubrak_design_2023} study the design and evaluation of a user interface for prescriptive process monitoring, highlighting a complementary concern with presenting actionable guidance to users. Our study does not evaluate explanation interfaces or human interaction; it examines the behavior of the underlying structural preferences.

We study this problem through automated musical harmonization. Melodies contain recurring and transposed motifs, while harmonies assign chords and functional relationships that may be preserved or changed across motif occurrences. This provides observable decision-level and relational outcomes. Computational models of tonality and music generation offer the domain foundation~\cite{chew_mathematical_2014,herremans_functional_2018}. Earlier studies examined quantum-inspired harmonic generation and maintainable hybrid architectures~\cite{pavlicek_quantum-inspired_2026,pavlicek_designing_2026}, alongside process-based visualization and analysis of harmony~\cite{almeida_bpmn_2024,pavlicek_integrating_2026,pavlicek_human-centered_2026,pavlicek_using_2026}. The present work instead isolates the effects of additional structural preferences under a fixed optimization procedure.

More broadly, process mining studies how recorded behavior can be related to process models~\cite{vander_aalst_process_2016}. A comparable methodological distinction matters here: matching individual decisions is not equivalent to preserving relationships between decisions. We therefore evaluate both chord-level agreement and context-stratified motif-pair relationships, without claiming that the musical results generalize directly to business processes.

We conduct a controlled empirical study of how structural preferences affect individual decisions and relationships between decisions in knowledge-guided generative systems. HARMONIX provides the experimental infrastructure: musical harmonization serves as a structured testbed for comparing recurring and transformed decision contexts. Four configurations---baseline (B), repetition (R), functional (F), and combined (RF)---share the same iterative Viterbi optimizer. We evaluate 19,745 input records and 143,634 motif pairs across development (TRAIN) and held-out validation (VALID) partitions. We address:
\begin{itemize}
\item \textbf{RQ1:} How do repetition and functional constraints affect overall agreement with human reference harmonizations?
\item \textbf{RQ2:} In which structural and transformation contexts do these constraints improve or reduce agreement with reference harmonic relationships?
\item \textbf{RQ3:} Can overall metrics hide opposing context-specific effects?
\end{itemize}

Our empirical contribution is threefold. First, a controlled ablation reveals opposing effects of structural preferences on reference root-interval agreement across contexts, reaching approximately +5 and $-9$ percentage points. Second, these directions persist across TRAIN and VALID and nine TRAIN weight settings. Third, near-unchanged aggregate chord agreement demonstrates why relational preferences require relational evaluation. This provides an empirical basis for assessing contextual constraint applicability separately from rule interpretability and aggregate performance, consistent with design-science evaluation~\cite{hevner_design_2004,gregor_positioning_2013}. Sections~2--8 present related work, the framework, experimental design, results, discussion, threats to validity, and conclusions.
\section{Background and Related Work}
\label{sec:related}
\subsection{Knowledge-Guided Generation and Structural Constraints}
Knowledge-guided generation represents domain knowledge as rules, constraints, or preferences. Hard constraints restrict feasible solutions; soft constraints influence their scores without excluding alternatives. Bonlarron et al.~\cite{bonlarron_constraining_2026} survey constraints in generative models. In information systems engineering, Corea et al.~\cite{corea_flexibility_2024} examine the flexibility of declarative process specifications, while Seidel et al.~\cite{seidel_recommendations_2024} study model-based recommendations for knowledge-intensive processes. These contributions motivate explicit decision guidance but do not imply that a particular relational preference is suitable in every context.

\subsection{Context and Constraint Applicability}
Bounded rationality and heuristic decision-making emphasize the dependence of decisions on available information and environmental conditions~\cite{simon_theories_1972,gigerenzer_heuristic_2011}. Similarly, repeated structures in reference data may suggest a useful preference without establishing its universal applicability. We distinguish the \emph{meaning} of a structural preference from its \emph{applicability}: whether rewarding that relationship increases agreement with an observed reference in a particular context. The reference is not assumed to be the only valid solution.

\subsection{Musical Harmonization as an Experimental Domain}
Harmonization assigns chords to melodic segments using local compatibility, harmonic transitions, and broader structure. Repeated motifs may retain or change their harmonic roles. Chew~\cite{chew_mathematical_2014} models tonal organization; Herremans et al.~\cite{herremans_functional_2018} classify music generation systems. Previous studies considered quantum-inspired harmonization and hybrid-system architecture~\cite{pavlicek_quantum-inspired_2026,pavlicek_designing_2026}; complementary studies examined BPMN and Chord Wheel representations~\cite{almeida_bpmn_2024,pavlicek_integrating_2026}. Here, the generator is held fixed to study the effects of structural preferences rather than competing architectures.

\subsection{Research Gap}
Constraint-based generation and explanation are well studied, but aggregate performance can conceal opposing effects on relationships between decisions. HARMONIX addresses this evaluation gap by comparing four configurations under an identical optimizer and examining both exact chord agreement and context-stratified reference root-interval agreement. The contribution is an empirical analysis of constraint applicability, not a new harmonization algorithm.

\section{The HARMONIX Framework}
\label{sec:framework}

\subsection{Framework Overview}

The system processes a melody as a sequence of
segments and assigns a chord to each segment.
The resulting chord assignments form a harmonic
sequence that is evaluated against a human-created
reference harmonization.

The processing pipeline consists of four main
components: input preparation, candidate generation,
optimization, and evaluation.

Input preparation converts musical data into
melodic segments. Candidate generation identifies
possible chords for each segment. The optimization
component selects a chord sequence based on local
musical scores and optional structural preferences.
Finally, the evaluation component compares the
generated sequence with reference harmonizations.

All four experimental configurations use the same
input representation, candidate generation, and
basic scoring mechanism. They differ only in the
additional structural preferences applied during
optimization. This controlled design allows us
to isolate and evaluate the effects of individual
constraints.

\begin{figure}[t]
    \centering
    \includegraphics[width=0.90\linewidth]{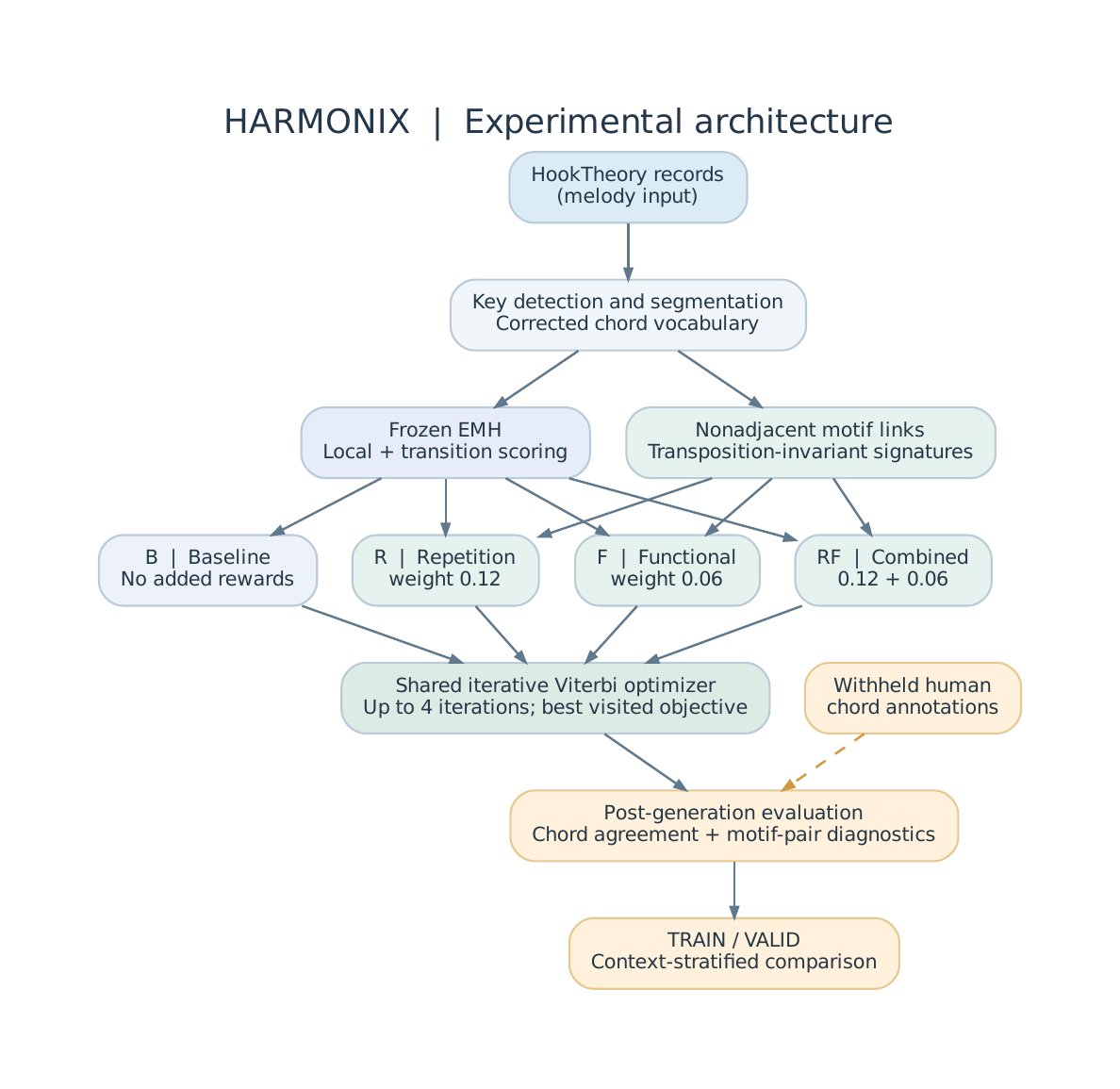}
    \caption{Architecture of the HARMONIX experimental
    framework. Four structural configurations share
    the same iterative optimization procedure.
    Human reference harmonizations are used only
    during post-generation evaluation.}
    \label{fig:harmonix_architecture}
\end{figure}

The architecture of the HARMONIX experimental
framework is illustrated in
Fig.~\ref{fig:harmonix_architecture}.

\subsection{Harmonic Decision Model}

Let a melody be represented by a sequence of
$T$ segments:

\begin{equation}
M = (m_1,m_2,\ldots,m_T).
\label{eq:melody}
\end{equation}

For each segment $m_t$, the system selects a chord
$c_t$ from a candidate set $\mathcal{C}$.
The complete harmonic solution is

\begin{equation}
H = (c_1,c_2,\ldots,c_T),
\qquad c_t \in \mathcal{C}.
\label{eq:harmony}
\end{equation}

Each chord is evaluated using several musical
properties. These include compatibility with
the melody, harmonic transition, chord persistence,
and cadential behavior.

We define the baseline objective as

\begin{equation}
S_0(H) =
\sum_{t=1}^{T}
s(m_t,c_{t-1},c_t),
\label{eq:baseline}
\end{equation}

where $s$ is the local scoring function.
For the first segment, the previous chord is
treated as an initial state.

The baseline optimizer selects a sequence that
maximizes this objective:

\begin{equation}
H_0^* =
\arg\max_{H \in \mathcal{H}} S_0(H),
\label{eq:baseline_opt}
\end{equation}

where $\mathcal{H}$ denotes the set of feasible
chord sequences.

The baseline objective contains local musical
knowledge. However, it does not explicitly reward
the preservation of harmonic relationships between
distant occurrences of repeated melodic motifs.

\subsection{Structural Constraints}

HARMONIX introduces two additional structural
preferences. Both are soft constraints. They
increase the score of certain solutions but do
not make other solutions automatically invalid.

First, the system identifies repeated melodic
patterns. Each segment receives a signature
based on relative pitch classes and normalized
note durations. Rests are represented explicitly.

Let $\mathcal{L}$ be the set of links between
non-adjacent occurrences of the same melodic
signature. The links connect consecutive
occurrences within each signature group.

\subsubsection{Repetition Constraint}

The repetition constraint rewards the use of
the same chord for linked melodic segments.

It is defined as

\begin{equation}
R(H) =
\sum_{(i,j)\in\mathcal{L}}
\mathbb{I}(c_i=c_j),
\label{eq:repetition}
\end{equation}

where $\mathbb{I}$ is an indicator function
equal to one when the condition is satisfied,
and zero otherwise.

This preference is based on the assumption
that repeated melodic structures may use
similar harmony. However, it does not mean
that all repeated motifs must have identical
chords.

\subsubsection{Functional Constraint}

The functional constraint considers relationships
between harmonic functions.

Let $f(c)$ denote the harmonic function
assigned to chord $c$. The functional
constraint rewards matching function pairs
around linked motif occurrences:

\begin{equation}
F(H) =
\sum_{\substack{(i,j)\in\mathcal{L}\\i>1}}
\mathbb{I}
\left(
f(c_{i-1})=f(c_{j-1})
\land
f(c_i)=f(c_j)
\right).
\label{eq:functional}
\end{equation}

This constraint compares the harmonic functions
at the linked positions and immediately before
them. It therefore considers a simple functional
transition rather than only the selected chord.

The two constraints describe different properties.
The repetition constraint concerns chord identity.
The functional constraint concerns relationships
between harmonic functions.

\subsection{Optimization and Experimental Configurations}

The structural objective combines the baseline
score with the two additional preferences:

\begin{equation}
S(H) =
S_0(H) +
\lambda_R R(H) +
\lambda_F F(H),
\label{eq:structural_objective}
\end{equation}

where $\lambda_R$ and $\lambda_F$ control
the influence of repetition and functional
constraints.

We evaluate four configurations (Table~\ref{tab:configurations}):

\begin{table}[t]
\centering
\caption{Experimental configurations of HARMONIX.}
\label{tab:configurations}
\begin{tabular}{lcc}
\hline
Configuration & $\lambda_R$ & $\lambda_F$ \\
\hline
Baseline & 0 & 0 \\
R        & 0.12 & 0 \\
F        & 0 & 0.06 \\
RF       & 0.12 & 0.06 \\
\hline
\end{tabular}
\end{table}

The baseline objective can be optimized using
Viterbi-style dynamic programming because its
score depends on the current and previous chord.

The structural objective is more difficult.
Links between repeated motifs can connect
positions that are far apart in the sequence.
These dependencies are not directly represented
by the ordinary Viterbi state.

HARMONIX therefore uses an iterative Viterbi
approximation. It first calculates the baseline
solution. In each following iteration, structural
bonuses are estimated using the chord assignments
from the previous iteration. A new chord sequence
is then calculated using dynamic programming.

The implementation uses four iterations as the
maximum. It also evaluates the complete structural
objective for the visited solutions and retains
the best one, including the initial baseline path.

This procedure is deterministic for a fixed
input and configuration. However, it does not
guarantee the global optimum of the complete
structural objective.

The same optimization procedure is used for
all structural configurations. Their differences
are determined by the two constraint weights.

Importantly, the optimization does not use
reference chord annotations. These are used
only during evaluation. This separation allows
us to study whether structural preferences
improve agreement with reference solutions
without directly optimizing against those
reference annotations.

The next section describes the datasets,
evaluation measures, and experimental procedure.

\section{Experimental Design}
\label{sec:experimental}

\subsection{Research Design}

We use a controlled empirical ablation to assess how structural preferences affect agreement with human-created reference harmonizations. Four configurations share the same inputs, candidate chords, baseline scores, and iterative Viterbi procedure: baseline (B), repetition (R), functional (F), and combined (RF). Each configuration generates a sequence for every record, enabling paired comparisons against the same reference. RQ1 addresses overall chord agreement, RQ2 context-dependent relational effects, and RQ3 whether aggregate evaluation conceals opposing effects.

\subsection{Dataset and Evaluation Units}

We retain the source dataset's TRAIN and VALID labels; HARMONIX does not perform supervised learning. TRAIN supported framework development and manual selection of structural weights. VALID was held out from parameter selection and evaluates the fixed configurations. The same scoring functions and optimizer are applied to both partitions. The labels thus distinguish development from evaluation, not model training.

The experiment uses chord positions for decision-level evaluation and links between nonadjacent occurrences of matching melodic signatures for relational evaluation (Section~3). Signatures encode relative pitch classes, normalized durations, and rests; links connect consecutive occurrences within each signature group. Motif pairs are not independent when they belong to the same record or musical work.

\subsection{Evaluation Measures}

We use exact chord agreement and reference root-interval agreement.

\subsubsection{Exact Chord Agreement}

Let $c_{u,t}^{(k)}$ and $c_{u,t}^{\mathrm{ref}}$ denote generated and reference chords at position $t$ in successfully processed record $u$, with $N_u$ chord positions. Agreement is macro-averaged over the $U$ successful records:
\begin{equation}
A_{\mathrm{chord}}^{(k)}=\frac{1}{U}\sum_{u=1}^{U}\left[\frac{1}{N_u}\sum_{t=1}^{N_u}\mathbb{I}\!\left(c_{u,t}^{(k)}=c_{u,t}^{\mathrm{ref}}\right)\right].
\label{eq:chord_accuracy}
\end{equation}
Each record contributes equally regardless of length.

\subsubsection{Reference Interval Agreement}

For a linked motif pair $(i,j)$, let $r(c)$ be the pitch class of chord $c$'s root. The directed root interval is
\begin{equation}
\Delta(c_i,c_j)=\left(r(c_j)-r(c_i)\right)\bmod 12.
\label{eq:root_interval}
\end{equation}
The generated relationship matches the reference if
\begin{equation}
\Delta(c_i^{(k)},c_j^{(k)})=\Delta(c_i^{\mathrm{ref}},c_j^{\mathrm{ref}}).
\label{eq:interval_match}
\end{equation}
For motif-pair set $\mathcal{P}$, pair-weighted agreement is
\begin{equation}
A_{\mathrm{int}}^{(k)}(\mathcal{P})=\frac{1}{|\mathcal{P}|}\sum_{(i,j)\in\mathcal{P}}\mathbb{I}\!\left(\Delta(c_i^{(k)},c_j^{(k)})=\Delta(c_i^{\mathrm{ref}},c_j^{\mathrm{ref}})\right).
\label{eq:interval_accuracy}
\end{equation}
This measure captures relationships between decisions even when individual generated chords differ from the reference.

\subsection{Structural Contexts and Comparison}

We stratify linked motif pairs by melodic transposition and the corresponding reference root relationship:
\begin{itemize}
\item \textbf{Untransposed, same root:} repeated melody without transposition; identical reference chord roots.
\item \textbf{Untransposed, changed root:} repeated melody without transposition; different reference roots.
\item \textbf{Transposed, tracking:} the reference root interval matches the melodic transposition.
\item \textbf{Transposed, non-tracking:} the reference root interval differs from the melodic transposition.
\end{itemize}
These labels depend partly on reference annotations and are used \emph{retrospectively for evaluation}, not to select constraints during generation. For configuration $k$ and context $\mathcal{P}_q$, the difference from B, in percentage points, is
\begin{equation}
D_{k,q}=100\left[A_{\mathrm{int}}^{(k)}(\mathcal{P}_q)-A_{\mathrm{int}}^{(B)}(\mathcal{P}_q)\right].
\label{eq:context_difference}
\end{equation}
Positive values indicate greater reference agreement; negative values indicate less. We report TRAIN and VALID separately.

\subsection{Uncertainty Estimation}

We estimate 95\% percentile confidence intervals for paired differences using 1,000 cluster-bootstrap resamples grouped by the available musical-work identifier, retaining all associated records and motif pairs within each sampled group. This grouping addresses within-work dependence to the extent that the available identifiers correctly identify works; it does not establish independence across groups. Exact chord agreement is recomputed as a record-level macro-average, while contextual root-interval agreement remains pair-weighted within each resample. The percentile context intervals are unadjusted for the 12 comparisons (three configurations across four contexts). For VALID, we additionally compute unstudentized simultaneous 95\% intervals using 5,000 paired work-proxy cluster-bootstrap resamples (seed 2027). The 95th percentile of the maximum absolute centered bootstrap error across all 12 contrasts gives a common half-width of 3.09 percentage points. These simultaneous intervals provide approximate family-wise coverage under the work-proxy clustering assumptions.

\subsection{Reproducibility and Research Artifacts}

The experimental implementation documents input processing, structural settings, optimization, evaluation, and bootstrap procedures. An anonymized research artifact may be supplied for double-blind review; a public repository reference can be restored after review.
\section{Results}
\label{sec:results}

\subsection{Overall Decision Agreement}
\label{sec:overall_results}

We first examine whether structural preferences change exact chord agreement
with the human reference harmonizations. We compare the baseline (B), repetition
preference (R), functional preference (F), and their combination (RF) on the
development (TRAIN) and held-out validation (VALID) partitions. Agreement is
calculated for each successfully processed record and then averaged across
records within each partition (macro-average). Thus, each record receives equal
weight, irrespective of its number of chord positions.

Table~\ref{tab:overall_chord_agreement} reports absolute agreement and the
changes relative to B. The analysis includes 17,911 successfully processed
TRAIN records and 1,821 VALID records; 12 TRAIN records and one VALID record
were excluded because processing was unsuccessful.

\begin{table}[t]
\centering
\caption{Mean exact chord agreement across records (\%) and changes relative
 to baseline B (percentage points, pp).}
\label{tab:overall_chord_agreement}
\small
\setlength{\tabcolsep}{4pt}
\begin{tabular}{lrrrr}
\hline
 & \multicolumn{2}{c}{TRAIN} & \multicolumn{2}{c}{VALID} \\
Configuration & Agreement & $\Delta$ (pp) & Agreement & $\Delta$ (pp) \\
\hline
B (baseline)   & 18.3225 & --      & 18.9664 & -- \\
R (repetition) & 18.4104 & +0.0879 & 19.0713 & +0.1049 \\
F (functional) & 18.3231 & +0.0006 & 19.0301 & +0.0636 \\
RF (combined)  & 18.3812 & +0.0587 & 19.0347 & +0.0683 \\
\hline
\end{tabular}
\end{table}

All aggregate differences are small. On VALID, R has the largest observed
increase (+0.1049 pp), followed by RF (+0.0683 pp) and F (+0.0636 pp).
The combination does not outperform R alone on this metric; the effects of
the two preferences are therefore not additive at the aggregate level.

The cluster-bootstrap 95\% confidence intervals for VALID differences relative
to B are $[-0.0639,\,0.2775]$ pp for R, $[-0.0400,\,0.1777]$ pp for F,
and $[-0.1174,\,0.2293]$ pp for RF. Each interval includes zero, so these
results do not establish an aggregate improvement in exact chord agreement.
Agreement with a single observed reference is not a measure of intrinsic
musical quality. Furthermore, aggregate agreement at individual chord
positions does not characterize relationships between distant decisions.

\subsection{Context-Specific Structural Effects}
\label{sec:context_results}

We next examine reference root-interval agreement for motif pairs in the four
contexts defined in Section~4.4. Unlike exact chord
agreement, this measure evaluates a relationship between decisions. For each
context, we report the change relative to B for R, F, and RF. Values are
percentage-point differences in pair-weighted agreement; positive values
indicate greater agreement with the observed reference root interval.
The TRAIN and VALID partitions contain 131,117 and 12,517 evaluable motif
pairs, respectively. VALID context sizes are 5,922, 2,639, 1,064, and
2,892 pairs in the order reported below.

Tables~\ref{tab:context_train} and~\ref{tab:context_valid} report the TRAIN and VALID ablations using identical contexts.

\begin{table}[t]
\centering
\caption{TRAIN: change in reference root-interval agreement relative to B
(percentage points).}
\label{tab:context_train}
\small
\setlength{\tabcolsep}{6pt}
\begin{tabular}{lrrr}
\hline
Context & R & F & RF \\
\hline
Untransposed, same root    & +4.78 & +3.87 & +5.08 \\
Untransposed, changed root & -1.44 & -0.73 & -1.52 \\
Transposed, tracking       & -8.08 & -1.59 & -8.80 \\
Transposed, non-tracking   & +3.95 & +0.74 & +4.13 \\
\hline
\end{tabular}
\end{table}

\begin{table}[t]
\centering
\caption{VALID: change in reference root-interval agreement relative to B
(percentage points).}
\label{tab:context_valid}
\small
\setlength{\tabcolsep}{6pt}
\begin{tabular}{lrrr}
\hline
Context & R & F & RF \\
\hline
Untransposed, same root    & +4.91 & +3.88 & +5.00 \\
Untransposed, changed root & -0.95 & -0.45 & -0.99 \\
Transposed, tracking       & -8.93 & -2.44 & -8.93 \\
Transposed, non-tracking   & +3.46 & +1.11 & +4.18 \\
\hline
\end{tabular}
\end{table}

For untransposed, same-root motifs, all preferences increase agreement; RF yields +5.08 pp on TRAIN and +5.00 pp on VALID. For changed-root motifs, all preferences reduce agreement.

For transposed tracking motifs, R reduces agreement by 8.08 pp on TRAIN and 8.93 pp on VALID, compared with 1.59 and 2.44 pp for F. RF reduces agreement by 8.80 and 8.93 pp, but improves non-tracking agreement by 4.13 and 4.18 pp, respectively.

R accounts for most of the tracking-context reduction. RF effects are not additive because configurations are optimized independently. Reference agreement does not establish musical quality.
For RF on VALID, the work-cluster bootstrap 95\% confidence intervals
(1,000 resamples) are [4.18, 5.96], [-1.70, -0.36], [-11.68, -6.45],
and [2.68, 5.68] pp, respectively, in the context order of
Table~\ref{tab:context_valid}. These intervals are unadjusted.
Across all 12 contrasts, the simultaneous 95\% intervals for RF on VALID are
[+1.91, +8.09], [-4.07, +2.10], [-12.02, -5.84], and [+1.10, +7.27] pp
in the same context order. Three of these four simultaneous intervals exclude zero;
only the untransposed changed-root contrast does not. The simultaneous
intervals use 5,000 work-proxy cluster-bootstrap resamples and a common
maximum-error half-width of 3.09 pp across all 12 comparisons.
Work-proxy clustering cannot guarantee cross-work independence.

\subsection{Consistency Across Data Partitions}
\label{sec:partition_consistency}

Across all four contexts, R, F, and RF have the same direction of change on
TRAIN and VALID. For RF, the absolute TRAIN--VALID differences in effect size
are 0.08 pp (untransposed, same root), 0.53 pp (untransposed, changed root),
0.13 pp (transposed, tracking), and 0.05 pp (transposed, non-tracking).

The TRAIN partition was used during framework development and manual weight
selection, whereas VALID was held out from this process. The similarity of
the context-dependent patterns on VALID therefore supports their consistency
within the evaluated dataset. The simultaneous intervals exclude zero for three of four RF contrasts on VALID; generalization to other musical corpora remains untested.

\subsection{Parameter Sensitivity}
\label{sec:parameter_sensitivity}

To examine dependence on the manually selected structural weights, we
re-evaluated TRAIN using a $3\times3$ grid:
$\lambda_R\in\{0.06,0.12,0.18\}$ and
$\lambda_F\in\{0.03,0.06,0.09\}$. The candidate generation, optimizer,
iteration limit, and evaluation contexts remained unchanged. Each setting
processed the same 17,923 input records, with 17,911 successful records and
131,117 evaluable motif pairs. This analysis was confined to TRAIN; the
prespecified configuration $(0.12,0.06)$ was retained for VALID.

For RF relative to B, the effect ranges across the nine settings were
$[+3.72,+7.05]$ pp (untransposed, same root),
$[-1.85,-1.16]$ pp (untransposed, changed root),
$[-9.00,-6.51]$ pp (transposed, tracking), and
$[+3.71,+4.26]$ pp (transposed, non-tracking).
All four effect directions were preserved at every setting (36 of 36
context--setting comparisons), although effect magnitudes varied. The
original RF setting produced $+5.08$, $-1.52$, $-8.80$, and $+4.13$ pp
on TRAIN, respectively. Thus, the observed directional trade-off is not
unique to one weight combination within the examined grid. These are
training-partition sensitivity results, not independent validation or
multiplicity-adjusted significance tests.

\subsection{Summary of Findings}
\label{sec:results_summary}

The results address the three research questions. For RQ1, all structural
configurations yield only marginal changes in overall exact chord agreement,
with VALID confidence intervals spanning zero. For RQ2, the ablation shows
that the effects of R, F, and RF depend on the reference relationship:
repetition-oriented guidance improves agreement in some contexts but reduces
it markedly when transposed motifs have tracking reference intervals.
For RQ3, the near-zero aggregate changes coexist with context-level changes
of approximately +5 pp and -9 pp for RF, with simultaneous 95\% intervals excluding zero for both of these VALID contrasts. The RF effect
directions also persist across nine TRAIN weight settings; their magnitudes
remain parameter-dependent.

Thus, evaluation at the level of individual decisions is insufficient to
characterize constraints that act on relationships between decisions.
The next section discusses the implications for knowledge-guided
generative systems.
\section{Discussion}
\label{sec:discussion}
\subsection{Structural Validity and Contextual Applicability}
The ablation distinguishes interpretable preferences from their contextual applicability. RF improves reference root-interval agreement for untransposed, same-root pairs by 5.08 pp on TRAIN and 5.00 pp on VALID, but reduces it for transposed, tracking pairs by 8.80 and 8.93 pp. Effects reverse in the other two contexts, showing that transposition alone does not determine applicability. R accounts for most of the tracking-context reduction, whereas F has a smaller effect. Because each configuration is optimized independently, RF effects are not additive. Thus, a meaningful structural preference need not be appropriate in every context~\cite{bonlarron_constraining_2026}; disagreement with one reference does not imply musical incorrectness.

\subsection{Why Aggregate Evaluation Is Insufficient}
RF changes macro-averaged exact chord agreement by only +0.0587 pp on TRAIN and +0.0683 pp on VALID, with a VALID confidence interval spanning zero. In contrast, context-stratified root-interval effects reach approximately +5 and -9 pp. These measures have different evaluation units, so this is not numerical cancellation. Rather, decision-level metrics alone fail to reveal changes in the relationships targeted by structural preferences.

\subsection{Implications for Information Systems Engineering}
Three design implications follow: define when each structural preference applies; evaluate relationships between decisions alongside decision accuracy, consistent with design science~\cite{hevner_design_2004,gregor_positioning_2013}; and distinguish explaining a rule from validating its contextual applicability~\cite{adadi_peeking_2018,barredo_arrieta_explainable_2020}.

HARMONIX makes these distinctions observable: chords are decisions, structural preferences are domain rules, and motif links represent relationships across recurring situations. Consider prescriptive process monitoring: a system may recommend the same action for similar cases even when changed operational conditions warrant different responses. Aggregate recommendation accuracy may remain stable while context-specific decision relationships deteriorate. Designers should therefore specify applicability conditions and test relational outcomes separately across contexts. This is a design implication illustrated by the musical experiment, not a cross-domain validation.

The contexts in our study are retrospective and partly reference-dependent; an operational context selector and explanation interfaces were not evaluated.

\subsection{Interpretation of the Two Data Partitions}
TRAIN supported weight selection; VALID used fixed settings. All 12 effects retain their directions across partitions, and all four RF directions persist across nine TRAIN settings (36/36 comparisons). This establishes stability only within the tested settings. VALID simultaneous 95\% intervals across 12 contrasts exclude zero for three RF contexts; cross-domain transfer remains untested.

\subsection{Research Artifacts and Reproducibility}
The HARMONIX implementation and experimental documentation support reproducibility. An anonymized research artifact can be supplied for double-blind review where the submission system permits; a public repository link may be released after review.

\section{Threats to Validity}
\label{sec:threats}
\textbf{Construct validity.} Agreement with one reference measures neither musical quality nor composer intent; alternative harmonizations may be valid. Retrospective context labels depend partly on reference roots, while root intervals omit voicing and expression.

\textbf{Internal validity.} Shared inputs, scoring, and optimization isolate the added preferences, but iterative Viterbi does not guarantee a global optimum. TRAIN-selected weights may introduce selection bias; the nine-setting sensitivity grid supports directional stability only within its tested range. VALID was not used for tuning.

\textbf{Statistical conclusion validity.} Chord agreement is record-macro-averaged, whereas contextual agreement is pair-weighted. VALID aggregate-effect intervals include zero. The original four RF context intervals use 1,000 work-cluster bootstrap resamples and are unadjusted; additional simultaneous intervals from 5,000 work-proxy cluster resamples exclude zero for three RF contrasts, not all four, across 12 comparisons. The sensitivity analysis is descriptive; its preliminary 100-resample intervals are not used for confirmatory inference. Work identifiers may imperfectly capture dependence. Of 19,745 records, 19,732 were processed successfully; bias from the remaining 13 failures cannot be excluded.

\textbf{External validity.} Only one corpus, chord vocabulary, motif construction, scoring model, and optimizer were studied. Generalization across genres or information systems domains and prospective context selection remain untested.

\section{Conclusion}
\label{sec:conclusion}
The central finding is that \emph{a constraint can be interpretable and locally reasonable yet systematically undermine reference relationships in specific contexts}. Across 19,745 records and 143,634 motif pairs, RF changed overall exact chord agreement only marginally, while shifting context-specific root-interval agreement by approximately +5 and -9 percentage points. Three of four RF context effects retain simultaneous 95\% intervals excluding zero across 12 contrasts on VALID. Thus, decision-level accuracy alone cannot evaluate rules that govern relationships between decisions. HARMONIX establishes a reproducible, two-level evaluation approach: assess individual decisions \emph{and} the context-dependent relationships targeted by structural preferences. For knowledge-guided information systems, contextual applicability must be evaluated separately from rule interpretability and aggregate performance.

\bibliographystyle{splncs04}
\section*{Code and Data Availability}
The HARMONIX source code, experimental configurations,
evaluation procedures, and available processed results
are publicly accessible in the project repository:

\begin{center}
\url{https://github.com/JosefPavlicek/quantum-inspired-music-research/}

\medskip
\textbf{GitHub -- HARMONIX}
\end{center}
The original HookTheory/SheetSage dataset is subject to
third-party licensing restrictions and cannot be
redistributed with the repository. Full reproduction of
the corpus-based experiments requires independent access
to the dataset from its authorized provider.
\bibliography{references_preprint}
\end{document}